%% file: WISENET2013.tex
\documentclass[conference]{IEEEtran}

\usepackage{graphicx}
\usepackage[cmex10]{amsmath}
\usepackage{algorithm,algorithmic}
\usepackage{amsfonts}
\usepackage{multirow}
\usepackage{color}
\usepackage{balance}

\input{Ahmed_defs}

\begin{document}
\title{\LARGE On Generalized Bayesian Data Fusion  with \\ Complex Models in Large Scale Networks}
\author{
\IEEEauthorblockN{Nisar Ahmed\IEEEauthorrefmark{1}, Tsung-Lin Yang\IEEEauthorrefmark{2}, and Mark Campbell\IEEEauthorrefmark{1}}
\IEEEauthorblockA{\IEEEauthorrefmark{1}Sibley School of Mechanical and Aerospace Engineering, \\ 
Cornell Unversity, Ithaca, NY 14853,  
email: (nra6, mc288)@cornell.edu}
\IEEEauthorblockA{\IEEEauthorrefmark{2}TaggPic,  
Ithaca, NY 14853,  
email: yangchuck@gmail.com}
}

\maketitle

\begin{abstract}
Recent advances in communications, mobile computing, and artificial intelligence have greatly expanded the application space of intelligent distributed sensor networks. This in turn motivates the development of generalized Bayesian decentralized data fusion (DDF) algorithms for robust and efficient information sharing among autonomous agents using probabilistic belief models. However, DDF is significantly challenging to implement for general real-world applications requiring the use of dynamic/ad hoc network topologies and complex belief models, such as Gaussian mixtures or hybrid Bayesian networks. To tackle these issues, we first discuss some new key mathematical insights about exact DDF and conservative approximations to DDF. These insights are then used to develop novel generalized DDF algorithms for complex beliefs based on mixture pdfs and conditional factors. Numerical examples motivated by multi-robot target search demonstrate that our methods lead to significantly better fusion results, and thus have great potential to enhance distributed intelligent reasoning in sensor networks. 
\end{abstract}

\IEEEpeerreviewmaketitle
\section{Introduction}
\input{intro}
\section{Background}
\input{background}
%
\section{Factorized DDF} \label{factoredddf}
\input{factoredddf}
\section{DDF with Mixture Model Factors} \label{hierarch}
\input{mixtureddf}

%
\bibliographystyle{IEEEtran}
\bibliography{WISENET2013_bib}
\end{document}

%% file: Ahmed_defs.tex
\newcommand{\eye}{\mbox{I}} 
\newcommand{\EV}[1]{\mathbb{E}\left[#1 \right]} 
\newcommand{\bigparenth}[1]{\left( #1\right)} 
\newcommand{\bigsquare}[1]{\left[ #1 \right]}
\newcommand{\inv}[1]{#1^{-1}} 

\newcommand{\set}[1]{\left\{ #1 \right\}}
\newcommand{\RealSpace}[1]{\mathbb{R}^{#1}}

\newcommand{\pareqref}[1]{(\ref{eq:#1})}

\newcommand{\smC}[1]{\mathrm{\textsc{#1}}} 

\newcommand{\pexactddf}[0]{p_{f}(x_k)}
\newcommand{\pwepddf}[0]{p_{f,\smC{WEP}}(x_k)}
\newcommand{\GaussianMix}[7]{\displaystyle \sum_{#2=1}^{#3^{#1}}{#4^{#1}_{#2}{\cal N}(#7;#5^{#1}_{#2},#6^{#1}_{#2})}} 
\newcommand{\wepcommoninfo}[2]{\hat{p}_c(#1;#2)}
\newcommand{\stategroup}[2]{\bar{x}^{#1}_{#2}}
\newcommand{\sg}[2]{\stategroup{#1}{#2}}
\newcommand{\KLD}[2]{\mbox{D}_{\smC{KL}}[#1 || #2]}

%% file: intro.tex
Intelligent robotic sensor networks have drawn considerable interest for applications like environmental monitoring, surveillance, search and rescue, and scientific exploration. To operate autonomously in the face of real world uncertainties, individual robots in such networks typically rely on perception algorithms rooted in Bayesian estimation methods \cite{ThrunBook}. These not only permit robots to make intelligent local decisions amid noisy data and complex dynamics, but also enable them to efficiently gather and share information with each other, which greatly improves perceptual robustness and task performance.

The Bayesian distributed data fusion (DDF) paradigm provides a particularly strong foundation for fully decentralized information sharing and perception in autonomous robot networks. In theory, DDF is mathematically equivalent to an idealized centralized Bayesian data fusion strategy (in which all raw sensor data is sent to a single location for maximum information extraction), but is far more computationally efficient, scalable, and robust to sensor network node failures through the use of recursive peer-to-peer message passing \cite{Grime94}. These properties have been successfully demonstrated for target search and tracking applications in large outdoor environments using wirelessly linked dynamic networks of autonomous/semi-autonomous UAVs \cite{Kaupp07,Ong08}.

Despite its merits, DDF is generally difficult to implement for three important reasons. Firstly, in order to maintain consistent network agent beliefs and avoid `rumor propagation' (i.e. double-counting old information as new information), either exact information pedigree tracking \cite{Martin05,Ong08} or conservative fusion \cite{Julier06} must be used. Each approach has different performance/robustness tradeoffs: exact pedigree tracking is optimal but computationally expensive for robust network communication topologies (i.e. loopy/ad hoc networks), whereas conservative fusion suboptimally loses some new information to ensure consistency under any topology. Secondly, both exact and conservative DDF methods yield analytically intractable results whenever local agent beliefs contain complex non-exponential family pdfs, e.g. Gaussian mixtures, which are commonly used for nonlinear estimation. Various approximations have been proposed to address this issue \cite{Julier06, Ong08, ChangSun10, Ahmed-RSS-12}, but these can be very inaccurate and do not scale well to large problem spaces. Thirdly, the DDF message passing protocol nominally requires each network node to exchange its entire local copy of the full joint state pdf with neighbors, which can lead to expensive processing requirements for high-dimensional pdfs in large/densely connected networks .

We propose novel solutions for the second and third issues. Specifically, we present new mathematical insights about exact and conservative DDF methods that lead to: (i) flexible `factorized DDF' updates, which greatly simplify communication and processing requirements for fusion with complex joint state pdfs, and (ii) accurate approximations for recursive fusion with finite mixture models over continuous random variables. Numerical examples in the context of large scale static target search show how our proposed methods lead to lower processing costs and more accurate fusion results compared to conventional DDF implementations. These results point to another interesting link between sensor networks and the powerful probabilistic graphical modeling framework (Bayes nets, MRFs, factor graphs, etc.); this can be exploited to develop novel tightly coupled perception and planning algorithms that enable decentralized mobile sensor networks to cope with complex uncertainties more efficiently and robustly.

%% file: background.tex
\subsection{Bayesian DDF Problem Formulation}
Let $x_k$ be a $d$-dimensional dynamic state vector of continuous and/or discrete random variables to be estimated by a decentralized network of $N_A$ autonomous agents. Assume each agent $i \in \set{1,...,N_A}$ can perform local recursive Bayesian updates on a common prior pdf $p_0(x)$ with sensor data $D^i_k$ having likelihood $p(D^i_k|x)$ at discrete time step $k>0$, so that 
\begin{align}
p_i(x|D^i_{1:k}) \propto p_i(x|D^i_{1:k-1}) \cdot p(D^i_k|x),
\label{eq:bayeslocal}
\end{align}
where $p_i(x|D^i_{1:k-1}) = p_0(x)$ for $k=1$. For brevity, the LHS of \pareqref{bayeslocal} is hereafter denoted as $p_i(x_k)$ (i.e. conditioning on available observations is always implied). Given an any node-to-node communication topology at $k$, assume $i$ is aware only of its connected neighbors and is unaware of the complete network topology. Let $N(i,k)$ denote the set of neighbors $i$ receives information \emph{from} at time $k$, and let $Z^i_k$ denote the set of information received by $i$ up to time $k$, i.e. $D^i_{1:k}$ and information previously sent to $i$ by other agents. The DDF problem is for each agent $i$ to find the fused information pdf
{
\allowdisplaybreaks
\abovedisplayskip = 2pt
\abovedisplayshortskip = 1pt
\belowdisplayskip = 2pt
\belowdisplayshortskip = 1pt
\begin{align}
p_f(x_k) \equiv p_i(x_k| Z^i_{k} \cup Z^{N(i,k)}_{k}).
\label{eq:mainFuse}
\end{align} 
}
Without loss of generality, assume $i$ computes (\ref{eq:mainFuse}) in a recursive `first in, first out' manner for each $j \in N(i,k)$. It is easy to show that $p_i(x_k) \propto p(x_k|Z^i_{k} \cap Z^j_k) p(x_k|Z^{i\slash j}_{k})$, where $Z^{i\slash j}_{k}$ is the exclusive information at agent $i$ with respect to $j$ and $p(x_k|Z^i_{k} \cap Z^j_k) \equiv p_c(x_k)$ is the \emph{common information pdf} between $i$ and $j$. Using this fact, \cite{Grime94} shows that \pareqref{mainFuse} can be exactly recovered via a distributed variant of Bayes' rule,
\begin{align}
\pexactddf = p_i(x_k | Z^i_{k} \cup Z^j_k) \propto \frac{p_i(x_k) p_j(x_k)}{p_c(x_k)}.
\label{eq:exactDef}
\end{align}
Note that $D^i_{1:k}$ and $D^j_{1:k}$ never need to be sent; $i$ and $j$ need only exchange their latest local pdfs for $x$, which compactly summarize all knowledge received from local sensor data and from network neighbors. To maintain consistency, \pareqref{exactDef} requires explicit tracking of $p_c(x_k)$, which can be handled through exact fusion algorithms like the channel filter \cite{Grime94, Ong08} or information graphs \cite{Martin05}. These methods are generally infeasible for dynamic ad hoc topologies, in which case suboptimal conservative approximations to (\ref{eq:exactDef}), such as the weighted exponential product (WEP) rule, can be used instead to guarantee consistent fusion without knowing $p_c(x_k)$, 
\begin{align}
\pwepddf \propto [p_i(x_k)]^{\omega} [p_j(x_k)]^{1-\omega},
\ \omega \in [0,1].
\label{eq:wepDef}
\end{align}
This trades off the amount of new information fused from $p_i(x_k)$ and $p_j(x_k)$ as a function of $\omega$, which can be optimized according to various convex information-theoretic cost metrics \cite{Julier06, Ahmed-RSS-12}. If only simple exponential family pdfs like Gaussians are required for estimation, then \pareqref{exactDef} and \pareqref{wepDef} always yield closed-form results that can be readily implemented via the exchange and manipulation of sufficient statistics.

\noindent
 Unfortunately, this is not the case for more complex pdfs such as Gaussian mixtures (GMs), which are widely used for nonlinear estimation applications. For example, consider a 2D Bayesian target search problem where $x \in \RealSpace{2}$ is the unknown location of a target in a large search space. As dicussed in \cite{FredThesis}, a decentralized team of mobile robots can use local optimal control laws in tandem with Bayesian DDF to efficiently reduce the uncertainty in $x$. 
Figure \ref{fig:fusion_demo} shows a simple example of a finite GM prior $p_0(x)$, along with a binary visual `detection/no detection' sensor model for an autonomous mobile robot and a finite GM approximation to \pareqref{bayeslocal}, 
{
\allowdisplaybreaks
\abovedisplayskip = 1pt
\abovedisplayshortskip = 1pt
\belowdisplayskip = 1pt
\belowdisplayshortskip = 0pt
\begin{align}
p_i(x | D^i_{1:k}) \approx \GaussianMix{i}{q}{M}{w}{\mu}{\Sigma}{x}
\label{eq:GMdef}
\end{align}
}
where $M^i$ is the number of mixands, $\mu^i_q$ and $\Sigma^i_q$ are mixand $q$'s mean and covariance matrix, and $w^i_q \in[0,1]$ is mixand $q$'s weight (s.t. $\sum_{q=1}^{M^i}{w^i_q}=1$). Since substitution of \pareqref{GMdef} into \pareqref{exactDef} or \pareqref{wepDef} leads to non-closed form fusion pdfs, we must find tractable yet accurate approximations to implement DDF with GMs. In particular, if $\pexactddf$ and $\pwepddf$ can always be closely approximated by GMs, then the recursive form of eqs. \pareqref{bayeslocal}, \pareqref{exactDef} and \pareqref{wepDef} can be (approximately) maintained. 

To this end, \cite{ChangSun10} derived a closed-form GM approximation to $\pexactddf$ that replaces the GM pdf $p_c(x_k)$ with a single moment-matched Gaussian, while \cite{Julier06} proposed a GM approximation to $\pwepddf$ that is based on the covariance intersection rule for Gaussian pdfs. Although fast and convenient, both methods rely on strong heuristic assumptions that lead to poor approximations of \pareqref{exactDef} and \pareqref{wepDef} whenever $p_i(x), p_j(x),$ or $p_c(x)$ are highly non-Gaussian. Refs. \cite{Ong08, Ahmed-RSS-12} proposed more rigorous approximations to \pareqref{exactDef} and \pareqref{wepDef} that use weighted Monte Carlo particle sets, which can be converted into GM pdfs via the expectation-maximization (EM) algorithm. However, these methods are computationally expensive for online operations, since they require a large number of particles and multiple EM initializations for robustness. Another concern is that the exchange of each agent's full local state pdf copy in \pareqref{exactDef} and \pareqref{wepDef} can lead to high communication and processing costs if either the state dimension is not fixed or if the full state pdf grows more complex over time due to nonlinear/non-Gaussian dynamics or observation models, e.g. as in Fig. \ref{fig:fusion_demo} (a) and (c). 
\input{gmcamerafusion}

%% file: gmcamerafusion.tex
\begin{figure}[t]
\centering
\begin{tabular}{@{}c@{}c@{}c@{}}
\includegraphics[width=3.0cm]{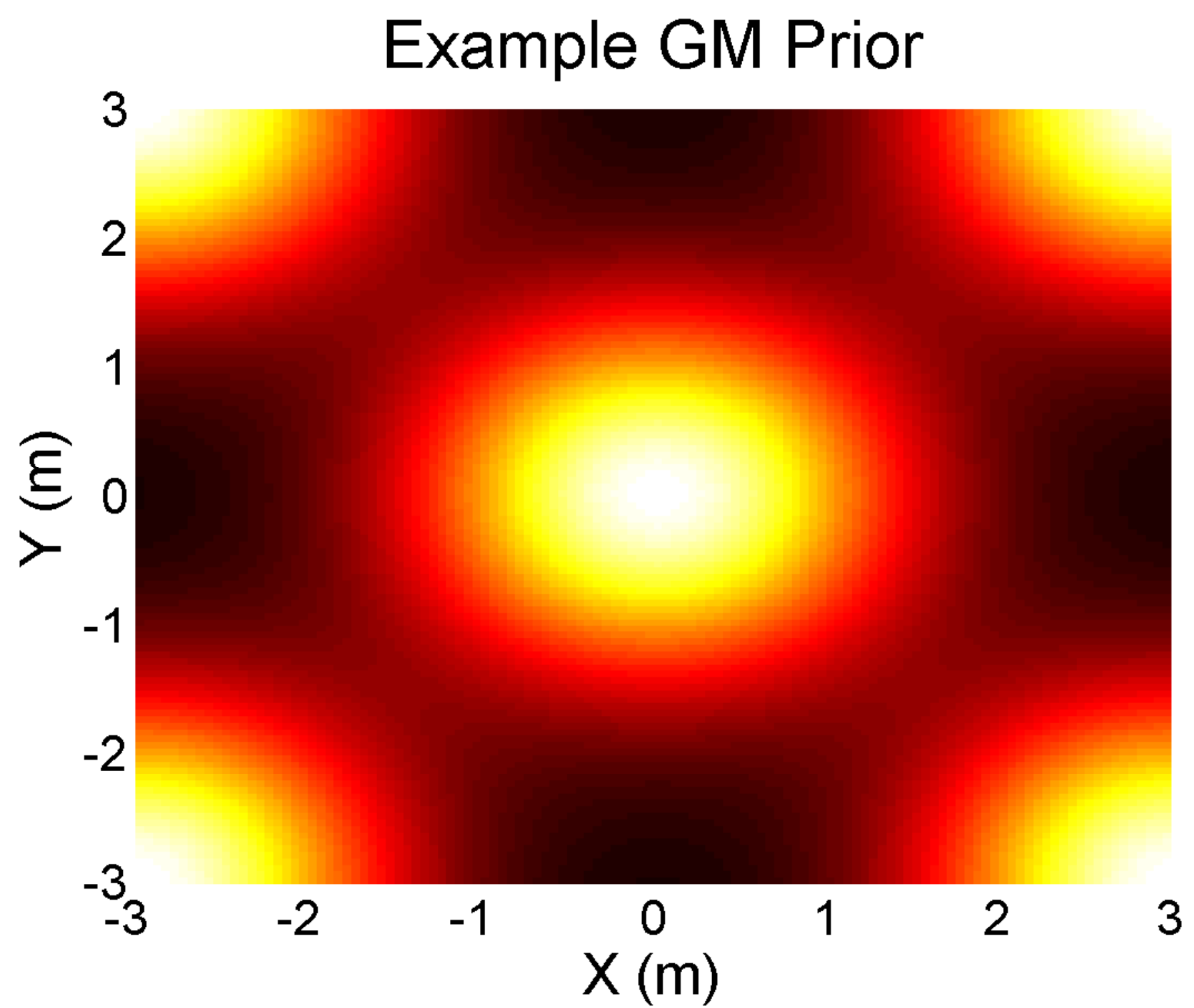} &
\includegraphics[width=3.0cm]{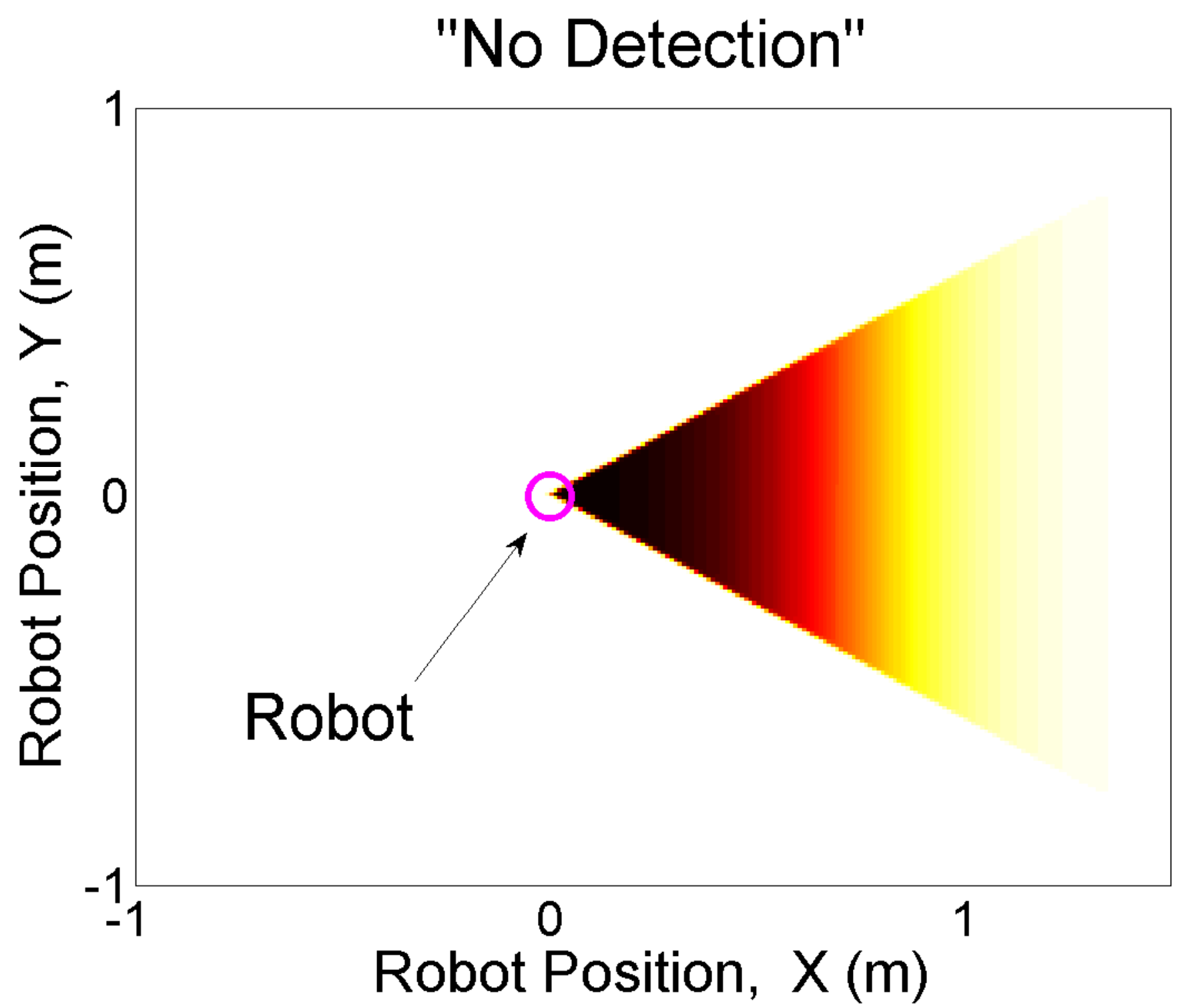} &
\includegraphics[width=3.0cm]{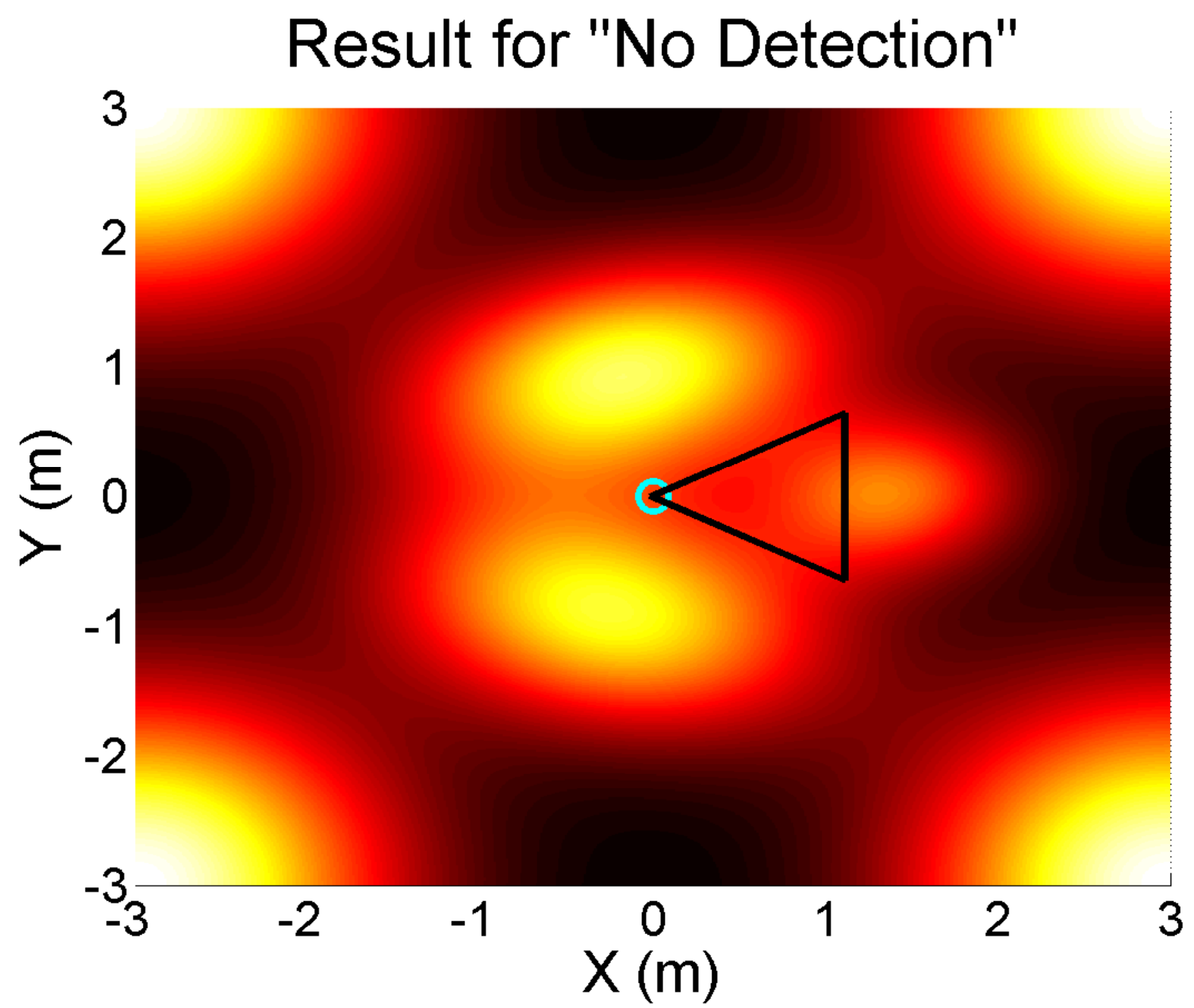} \\
\scriptsize (a) & \scriptsize (b) & \scriptsize (c)
\end{tabular}
\caption{Bayesian GM fusion example (black/white = low/high probability; magenta circle shows robot's position): (a) prior GM pdf $p_0(x)$, (b) binary visual detector model $p(D|x)$, (c) Bayesian posterior GM pdf $p(x|D)$.}
\label{fig:fusion_demo}
\vspace{-0.1in}
\end{figure}

%% file: factoredddf.tex
We propose a novel way to implement eqs. \pareqref{exactDef} and \pareqref{wepDef} that allows agents to selectively exchange \emph{partial copies} of complex state pdfs $p_i(x_k)$ and $p_j(x_k)$, so that relevant new information about different subsets of $x$ can be shared more efficiently. This is accomplished by rewriting \pareqref{exactDef} and \pareqref{wepDef} in terms of conditional dependencies within $x$, so that $p_i(x_k)$ and $p_j(x_k)$ factor into smaller conditional pdfs that are easier to communicate and process. 
\subsection{Factorized Exact DDF}
\noindent
Suppose $x_k = [x^1_k,...,x^d_k]$; let $\stategroup{s}{k}$ represent an arbitrary grouping of sub-states of $x_k$, such that $\bigcup_{s}\stategroup{s}{k} = x_k \forall s\in \set{1,...,N_s}$ and $\stategroup{s_1}{k} \bigcap \stategroup{s_2}{k} = \emptyset, \forall s_1 \neq s_2$ (the ordering of states in each grouping is unimportant, but each $\stategroup{s}{k}$ has at least 1 state).  
Then the law of total probability implies that
{
\allowdisplaybreaks
\abovedisplayskip = 2pt
\abovedisplayshortskip = 2pt
\belowdisplayskip = 2pt
\belowdisplayshortskip = 0pt
\begin{align}
p_i(x_k) &= p_i(\sg{1}{k}|\sg{2}{k},...,\sg{N_s}{k}) p_i(\sg{2}{k}|\sg{3}{k},...,\sg{N_s}{k}) ... p_i(\sg{N_s}{k}) \nonumber \\
&= \bigparenth{\prod_{s=1}^{N_s-1}{p_i(\sg{s}{k}|\sg{s+1:N_s}{k}) } }  p_i(\sg{N_s}{k}), \nonumber
\end{align}
}
where all terms are implicitly conditioned on $Z^i_k$. Applying this factorization to \pareqref{exactDef} gives
{
\allowdisplaybreaks
\abovedisplayskip = 2pt
\abovedisplayshortskip = 2pt
\belowdisplayskip = 2pt
\belowdisplayshortskip = 0pt
\begin{align}
&p_f(x_k) \propto \nonumber 
&\frac{ \prod_{w\in \set{i,j}} \bigparenth{\prod_{s=1}^{N_s-1}{p_w(\sg{s}{k}|\sg{s+1:N_s}{k}) } } p_w(\sg{N_s}{k})} {\bigparenth{\prod_{s=1}^{N_s-1}{p_c(\sg{s}{k}|\sg{s+1:N_s}{k}) } } p_c(\sg{N_s}{k})}. \nonumber
\end{align}
}
Grouping like terms together and simplifying yields
{
\allowdisplaybreaks
\abovedisplayskip = 2pt
\abovedisplayshortskip = 2pt
\belowdisplayskip = 2pt
\belowdisplayshortskip = 0pt
\begin{align}
p_f(x_k) &\propto 
\bigparenth{\prod_{s=1}^{N_s-1}{\frac{p_i(\sg{s}{k}|\sg{s+1:N_s}{k})  p_j(\sg{s}{k}|\sg{s+1:N_s}{k}) }{p_c(\sg{s}{k}|\sg{s+1:N_s}{k})} \eta(\sg{1:s}{k}) }}\nonumber \\  
&\times \frac{p_i(\sg{N_s}{k}) p_j(\sg{N_s}{k})}{p_c(\sg{N_s}{k})},
\label{eq:genfactorexactDDF}
\end{align}
}
Thus, the original DDF update for a single $d$-dimensional joint pdf is equivalent to $N_s \leq d$ separate \emph{conditional} DDF updates. This means that agents $i$ and $j$ can exchange information about various `chunks' of the state pdf, e.g. the latest pdfs for $(x_2,x_3|x_4,x_5...,x_d)$ and $(x_{d-1}|x_d$) may be fused for certain values of $x_4,x_5...,x_d$ during one exchange, while other conditional factors are fused during other exchanges.  
\subsection{Factorized WEP DDF}
The factorization principle extends to approximate WEP DDF for dynamic ad hoc network topologies, where exact tracking and removal of $p_c(x_k)$ is infeasible. This follows from the fact that eq. \pareqref{wepDef} can be rewritten as 
\begin{align}
\pwepddf &\propto \frac{p_i(x_k)p_j(x_k)}{[p_i(x_k)]^{1-\omega}[p_j(x_k)]^{\omega}} 
\propto \frac{p_i(x_k)p_j(x_k)}{\wepcommoninfo{x_k}{\omega}},
\label{eq:wepcommoninfo}
\end{align}
where $\wepcommoninfo{x_k}{\omega} \propto [p_i(x_k)]^{1-\omega}[p_j(x_k)]^{\omega}$ can be thought of as a conservative estimate of the common information pdf. This simple yet novel insight allows to write, as in eq. \pareqref{genfactorexactDDF},
{
\allowdisplaybreaks
\abovedisplayskip = 2pt
\abovedisplayshortskip = 2pt
\belowdisplayskip = 2pt
\belowdisplayshortskip = 0pt
\begin{align}
&\pwepddf \propto \nonumber \\
&=\bigparenth{\prod_{s=1}^{N_s-1}{\frac{p_i(\sg{s}{k}|\sg{s+1:N_s}{k})  p_j(\sg{s}{k}|\sg{s+1:N_s}{k}) }{\wepcommoninfo{\sg{s}{k}|\sg{s+1:N_s}{k}}{\omega}}}}  \cdot \frac{p_i(\sg{N_s}{k}) p_j(\sg{N_s}{k})}{\wepcommoninfo{\sg{N_s}{k}}{\omega}},
\label{eq:genfactorwepDDF}
\end{align}
} 
where each denominator term is a conservative estimate of a \emph{conditional} common information pdf. Note that \pareqref{wepcommoninfo} and \pareqref{genfactorwepDDF} nominally imply that these terms share the same $\omega$, since
{
\allowdisplaybreaks
\abovedisplayskip = 2pt
\abovedisplayshortskip = 2pt
\belowdisplayskip = 2pt
\belowdisplayshortskip = 0pt
\begin{align}
&\hat{p}_c(x^1_k,...,x^d_k) \propto [p_i(\sg{1}{k},...,\sg{N_s}{k})]^{1-\omega}[p_j(\sg{1}{k},...,\sg{N_s}{k})]^{\omega} \nonumber \\
&\propto \prod_{s=1}^{N_s-1}{[p_i(\sg{s}{k}|\sg{s+1:N_s}{k})]^{1-\omega}  [p_j(\sg{s}{k}|\sg{s+1:N_s}{k})]^{\omega}} \nonumber \\
&\ \ \ \ \ \ \ \ \ \times [p_i(\sg{N_s}{k})]^{1-\omega} [p_j(\sg{N_s}{k})]^{\omega}
\end{align}
}
However, it is possible to specify a separate $\omega$ parameter for each estimated conditional common information term, i.e.
{ 
\allowdisplaybreaks
\abovedisplayskip = 2pt
\abovedisplayshortskip = 2pt
\belowdisplayskip = 2pt
\belowdisplayshortskip = 0pt
\begin{align}
&\hat{p}_c(x^1_k,...,x^d_k) \nonumber \\
&\propto \prod_{s=1}^{N_s-1}{[p_i(\sg{s}{k}|\sg{s+1:N_s}{k})]^{1-\omega_{s|s+1:N_s}}  [p_j(\sg{s}{k}|\sg{s+1:N_s}{k})]^{\omega_{s|s+1:N_s}}} \nonumber \\ 
& \ \ \ \ \ \ \ \ \times [p_i(\sg{N_s}{k})]^{1-\omega_d} [p_j(\sg{N_s}{k})]^{\omega_d} \nonumber \\
&\propto \prod_{s=1}^{N_s-1}{\wepcommoninfo{\sg{s}{k}|\sg{s+1:N_s}{k}}{\omega_{s|s+1:N_s}}} \cdot \wepcommoninfo{\sg{N_s}{k}}{\omega_{N_s}},  
\end{align}
}
so that the WEP update can be generally expressed as
{
\allowdisplaybreaks
\abovedisplayskip = 2pt
\abovedisplayshortskip = 2pt
\belowdisplayskip = 2pt
\belowdisplayshortskip = 0pt
\begin{align}
&\pwepddf \propto \nonumber \\
&\bigparenth{\prod_{s=1}^{N_s-1}{\frac{p_i(\sg{s}{k}|\sg{s+1:N_s}{k})  p_j(\sg{s}{k}|\sg{s+1:N_s}{k}) }{\wepcommoninfo{\sg{s}{k}|\sg{s+1:N_s}{k}}{\omega_{s|s+1:N_s}}}}} \nonumber \\ 
&\times \frac{p_i(\sg{N_s}{k}) p_j(\sg{N_s}{k})}{\wepcommoninfo{\sg{N_s}{k}}{\omega_{N_s}}}.
\label{eq:genfactorwepDDF}
\end{align}
} 
The parameters $\omega_{s|s+1:N_s}$ and $\omega_{N_s}$ can be separately optimized using the information-theoretic cost metrics described in \cite{Ahmed-RSS-12, Julier06}. However, the main advantage of \pareqref{genfactorwepDDF} lies in the fact that the dependence of $\omega_{s|s+1:N_s}$ on $x_{s+1:N_s}$ can also be exploited to further minimize information loss for each conditional state $x_{s|s+1:N_s}$.
\subsection{Exploiting Conditional Independence}
To obtain a complete state update in \pareqref{genfactorexactDDF} or \pareqref{genfactorwepDDF}, each factor must be evaluated with respect to all possible configurations of up to $d-1$ conditioning states. This can be computationally expensive/intractable for large $d$, especially if any conditioning states are continuous or are discrete with many possible realizations. This issue can be greatly mitigated by exploiting conditional independence relationships among the state groupings $\sg{s}{k}$. For instance, if $x_k = [\sg{\#}{k}, \sg{*}{k}]$, where $\sg{\#}{k}$ is a partition of states that are conditionally independent of each other given another smaller partition of states $\sg{*}{k}$ and sensor data, then
{
\allowdisplaybreaks
\abovedisplayskip = 2pt
\abovedisplayshortskip = 2pt
\belowdisplayskip = 2pt
\belowdisplayshortskip = 0pt
\begin{align}
p_f(x_k) &\propto \bigparenth{\prod_{x^s_k \in \sg{\#}{k}}{ \frac{p_i(x^s_k|\sg{*}{k})  p_j(x^s_k|\sg{*}{k}) } {p_c(x^s_k|\sg{*}{k})}} \eta(x^s_k,\sg{*}{k})} \nonumber \\ 
&\times \frac{p_i(\sg{*}{k}) p_j(\sg{*}{k})}{p_c(\sg{*}{k})}\eta(\sg{*}{k}),
\label{eq:markovblanketDDF}
\end{align}
} 
\noindent
where updates for $x^s_k \in \sg{\#}{k}$ all depend on the same fixed number of states in $\sg{*}{k}$. 
In many cases, $x_k$ can also be augmented with latent variables to introduce useful conditional factorizations that lead to more efficient processing.
In general, this implies that factorized DDF can be quite useful whenever the local posteriors $p_i(x_k)$ and $p_j(x_k)$ can be represented via modular/hierarchical factors, such as those used in probabilistic graphical models like undirected Markov random fields (MRFs) or hybrid directed Bayesian networks (BNs) \cite{Bishop06book}. 

Although a full mathematical treatment is beyond the scope of this note, the following target search example gives a simple illustration of how intelligent sensor agents can leverage such probabilistic graphical models to manage and share complex hybrid information efficiently and flexibly via factorized DDF.
\subsection{Target Search Example with Hybrid State Model}
\input{hybridBNSearchSetup}
\input{searchResults}
\input{omegaRsens}
Figure \ref{fig:hybridBN} (a) shows the physical setup for a search problem in which multiple mobile robots are looking for a static object in a large open space (Cornell's Engineering Quadrangle). Here, target coordinates $x^1$ and $x^2$ are grouped together into the 2D random variable $x$, which in turn is partitioned into $N_R \geq1$ mutually exclusive discrete regions by latent random variable $R$; each $R \in \set{1,...,N_R}$ is assigned a prior pdf $p_0(x|R)$ and region probability $p_0(R) \in [0,1]$ s.t. $\sum_{R}{p_0(R)=1}$. Let $D^i_k$ be conditionally dependent on $R$ such that the target is only detectable in $R$ if that region is in the robot's sensor range, i.e.
{
\allowdisplaybreaks
\abovedisplayskip = 2pt
\abovedisplayshortskip = 2pt
\belowdisplayskip = 2pt
\belowdisplayshortskip = 0pt
\begin{align}
&p(D^i_k = \mbox{`no detection'}| x, R=r) = 1, \ \mbox{if in $r$ sensor range} \nonumber \\
&p(D^i_k | x, R=r) = p(D^i_k|x) \ \mbox{from Fig. 1(b), otherwise}. \nonumber
\end{align}
}
Figure \ref{fig:hybridBN} (b) shows the corresponding hybrid BN model used by each robot to update its local belief over $x$ and $R$. Considering local updates for robot $i$ (likewise for $j$), the joint pdf from the hybrid BN is
{
\allowdisplaybreaks
\abovedisplayskip = 2pt
\abovedisplayshortskip = 2pt
\belowdisplayskip = 2pt
\belowdisplayshortskip = 0pt
\begin{align}
p_i(x, R, D^i_{1:k}) = p_0(R)p_0(x|R)p(D^i_{1:k}|x, R), \nonumber 
\end{align}
}
and so the Bayesian sensor update can be factored as
{
\allowdisplaybreaks
\abovedisplayskip = 2pt
\abovedisplayshortskip = 2pt
\belowdisplayskip = 2pt
\belowdisplayshortskip = 0pt
\begin{align}
&p_i(x, R |D^i_{1:k}) = p_i(x|R,D^i_{1:k}) \cdot  p_i(R|D^i_{1:k}).
\label{eq:localfactorization}
\end{align}
}
The conditional pdfs are recursively updated via Bayes' rule,
{
\allowdisplaybreaks
\abovedisplayskip = 2pt
\abovedisplayshortskip = 2pt
\belowdisplayskip = 2pt
\belowdisplayshortskip = 0pt
\begin{align}
&p_i(x|R,D^i_{1:k}) \propto p_i(x|R,D^i_{1:k-1}) p_i(D^i_{k}|x,R), \label{eq:localxupdate} \\
&p_i(R|D^i_{1:k}) \propto p_i(R|D^i_{1:k-1}) p(D^i_k|R,D^i_{1:k-1}) , \label{eq:localRupdate}
\end{align}
}
where 
$p(D^i_k|R,D^i_{1:k-1}) = \int{p_i(x|R,D^i_{1:k-1}) p_i(D^i_{k}|x,R)  dx}$. In this hybrid model, each robot updates only its local copy of $p(x|R)$ if $R$ is within sensor range. The robots can then selectively fuse posterior regional pdfs $p(x|R)$ and/or the whole set of posterior discrete region weights $p(R)$ with each other via either factorized exact or WEP DDF, 
\begin{align}
p_f(x,R) &\propto \frac{p_i(x|R) p_j(x|R)}{p_c(x|R)} \cdot \frac{p_i(R) p_j(R)}{p_c(R)}  \nonumber \\
&= p_f(x|R) \cdot p_f(R) \cdot \eta(R), \label{exactFacXR} \\
&p_{f,\smC{WEP}(x,R)} \propto \frac{p_i(x|R) p_j(x|R)}{\hat{p}_c(x|R; \omega_{x|R})} \cdot \frac{p_i(R)p_j(R)}{\hat{p}_c(R; \omega_R)}  \nonumber \\
&= p_{f,\smC{WEP}}(x|R; \omega_{x|R}) \cdot p_{f,\smC{WEP}}(R;\omega_R) \cdot \hat{\eta}(R), \label{wepFacXR}
\end{align}    
where $p_f(\cdot)$ and $p_{f,\smC{WEP}}(\cdot)$ refer to \emph{locally normalized} conditional fusion posteriors, and $\eta(R)=\int{\frac{p_i(x|R)p_j(x|R)}{p_c(x|R)}dx}$ and $\hat{\eta}(R) =\int{\frac{p_i(x|R)p_j(x|R)}{\hat{p}_c(x|R;\omega)}dx}$ are the `denormalization' terms required to make the product of the normalized conditional fusion posteriors equal to their corresponding unnormalized joint fusion pdfs. 

As a simple numerical example, consider a search mission for two robot agents who are initialized with the same prior search map shown in Figure \ref{fig:searchResults} (a), where $p_0(x|R)$ is given by a discrete grid approximation to a pseudo-uniform GM pdf for each $R=r \in \set{1,...,6}$ and $p_0(R) = [0.1190, 0.1190, 0.2415, 0.1497, 0.1735, 0.1973]$. For time steps $k=0$ to $k=600$, robot 1 starts off in region $R=5$ and moves in a counterclockwise inward spiral through regions $R=2, 1, 4$ and $5$, while robot 2 starts off in region $R=3$ and moves in a counterclockwise inward spiral through regions $R= 2, 4, 6,$ and $3$ while locally fusing its own sensor data. The robots perform Bayesian sensor updates using only their own local data up time step $k=600$, at which point they decide to perform a DDF update. Note that only robot 1 has any new information about $R \in \set{1,4}$ and only robot 2 has any new information about $R \in \set{3,5}$, while both robots $1$ and $2$ have new information about $R \in \set{2,5}$. It is thus straightforward to show that factorized exact DDF gives  
\begin{align}
&p_f(x|R \in \set{1,4}) = p_1(x|R\in \set{1,4}), \nonumber \\
&p_f(x|R \in \set{3,6}) = p_2(x|R\in \set{3,6}), \nonumber \\
&p_f(x|R \in \set{2,5}) \propto \frac{p_1(x|R\in \set{2,5}) p_2(x|R\in \set{2,5})}{p_0(x|R\in \set{2,5})}, \nonumber \\
&p_f(R) \propto \frac{p_1(R|D^1_{1:600}) \cdot p_j(r|D^2_{1:600})}{p_0(R)},  \nonumber \\
&\propto p_0(R) p_1(D^1_{1:600}|R) p_2(D^2_{1:600}|R), \ \forall R. \nonumber
\end{align}
This implies that robot 1 needs to send $p(R)$ (or $p_1(D^1_{1:600}|R\in \set{1,2,4,5})$) and $p(x|R\in \set{1,2,4,5})$ to robot 2, which in turn only needs to send $p(R)$ (or $p_2(D^2_{1:600}|R\in \set{2,3,4,6})$) to robot 1. Once robot 2 receives robot 1's message, it directly overwrites its local copy of $p_2(x|R\in \set{1,4})$ with $p_f(x|R\in \set{1,4}) = p_1(x|R\in \set{1,4})$, calculates $p_f(x|R \in \set{2,5})$ as above, and finally updates $p(R)$, while leaving $p_2(x|R\in \set{3,6})$ unaltered. Robot 1 performs a similar update procedure upon receiving robot 2's message, except that it directly overwrites its local copy of $p_1(x|R\in \set{3,6})$ with $p_2(x|R\in \set{3,6})$ and leaves $p_1(x|R\in \set{1,4})$ unaltered. In this manner, each robot performs a full hybrid state pdf update with a sparse set of messages and calculations that exactly recovers the centralized fusion pdf as shown by the final grid-based result in Figure \ref{fig:searchResults} (b), thus bypassing the need to transmit/process the raw sensor data histories or each robot's full local copy of $p(x,R|D_{1:600})$. 

Similar communication and processing requirements are obtained for factorized WEP DDF, if robots 1 and 2 directly set $p_{f,\smC{WEP}}(x|R; \omega_{x|R})$ for $R\in \set{1,3,4,6}$ using
\begin{align}
\omega(x|R \in \set{1,4}) &= 1 \mbox{ \ (robot 1) /} \ 0 \mbox{ \ (robot 2)}, \nonumber \\
\omega(x|R \in \set{3,6}) &= 0 \mbox{ \ (robot 1) /} \ 1 \mbox{ \ (robot 2)}, \nonumber \\
\Rightarrow p_{f,\smC{WEP}}(x|R \in \set{1,4}) &= p_1(x|R\in \set{1,4}), \nonumber \\
p_{f,\smC{WEP}}(x|R \in \set{3,6}) &= p_2(x|R\in \set{3,6}), \nonumber
\end{align}
and use the minimax WEP metric described in \cite{Ahmed-RSS-12} to perform three separate optimizations (two for $\omega_{x|R}$ for $R\in \set{2,5}$ and one for $\omega_R$). However, Figure \ref{fig:searchResults}(c) shows that the final fusion result incurs an information loss of 0.0214 nats, as given by the joint Kullback-Leibler divergence (KLD)
\begin{align}
&\KLD{p_f(x,R)}{p_{f,\smC{WEP}}(x,R)} = \nonumber \\ 
&\KLD{p_f(R)}{p_{f,\smC{WEP}}(R)} + \sum_{r\in R}{p(r)\KLD{p_{f}(x|r)}{p_{f,\smC{WEP}}(x|r)}},  \nonumber
\end{align}
where
\begin{align}
&\KLD{p_f(R)}{p_{f,\smC{WEP}}(R)} = \sum_{r\in R}{p(R)\log \frac{p_f(r)}{p_{f,\smC{WEP}}(r)}}, \nonumber \\
&\KLD{p_f(x|r)}{p_{f,\smC{WEP}}(x|r)} = \int{p_f(x|r)\log \frac{p_f(x|r)}{p_{f,\smC{WEP}}(x|r)} dx}. \nonumber
\end{align} 
Closer inspection reveals that $\KLD{p_f(R)}{p_{f,\smC{WEP}}(R)}$ contributes the most to joint KLD, while $\KLD{p_f(x|R)}{p_{f,\smC{WEP}}(x|R)}$ is extremely small for $R \in \set{2,5}$ and zero for $R \in \set{1,3,4,6}$. 
Interestingly, Fig. \ref{fig:omegaRsens} shows that this information loss can be minimized by choosing an $\omega_R$ value larger than the one found via the minimax WEP metric (where all $\omega_{x|R}$ are held fixed). Fig. \ref{fig:omegaRsens} also shows this new $\omega_R$ leads to lower information losses than application of conventional `whole WEP' DDF over the joint 3-dimensional grid for $x$ and $R$ (i.e. the same as setting all $\omega_{x|R} = \omega_R$). While these results imply that factorized WEP can indeed lead to very accurate fusion results, more sophisticated procedures for joint optimization of $\omega_R$ and $\omega_{x|R}$ must be found to minimize information loss. 

%% file: hybridBNSearchSetup.tex
\begin{figure}[t]
\centering
\begin{tabular}{@{}c@{}c@{}}
\hspace*{-0.1 in}
\includegraphics[width=5.3cm]{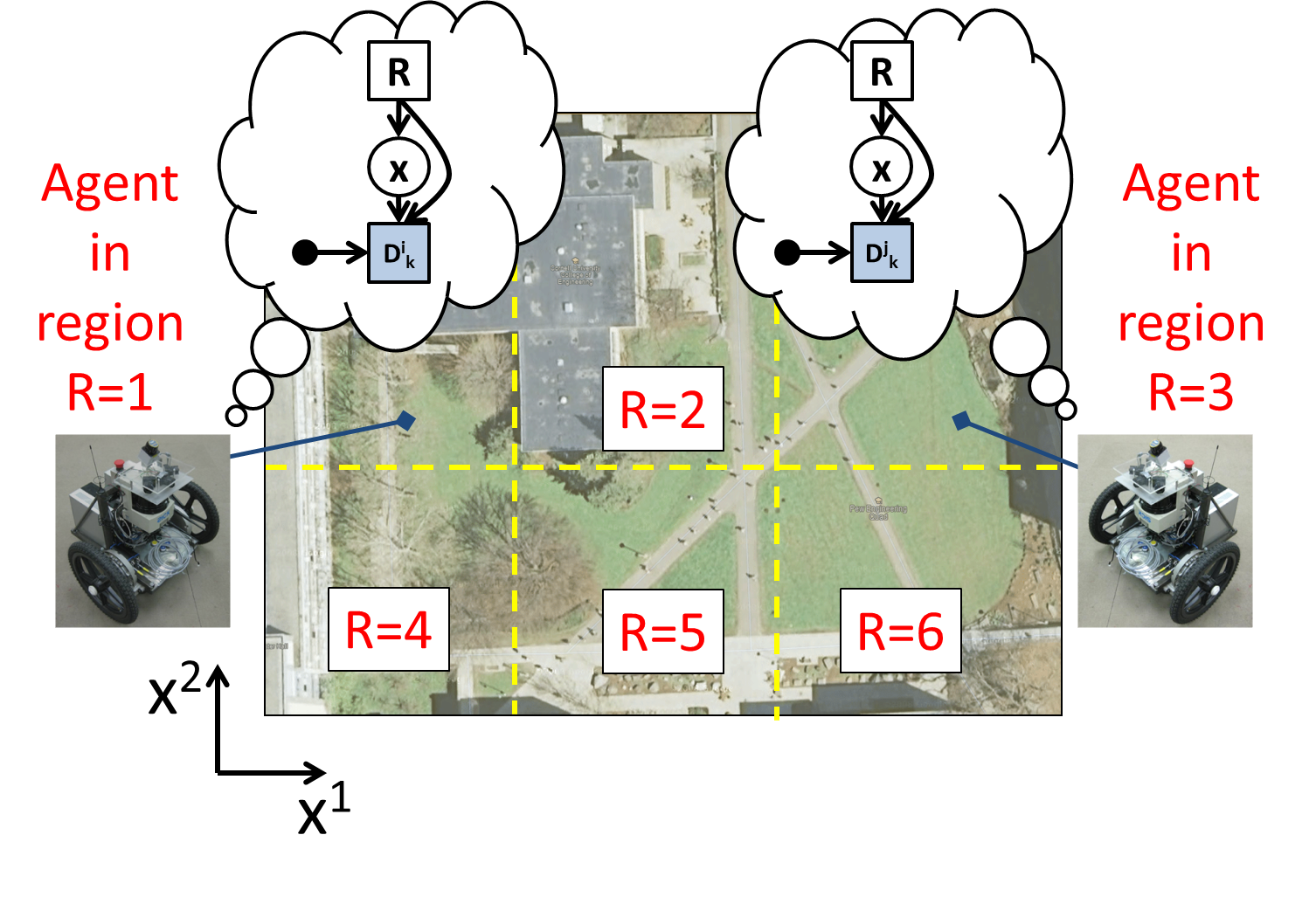} &
\includegraphics[width=3.5cm]{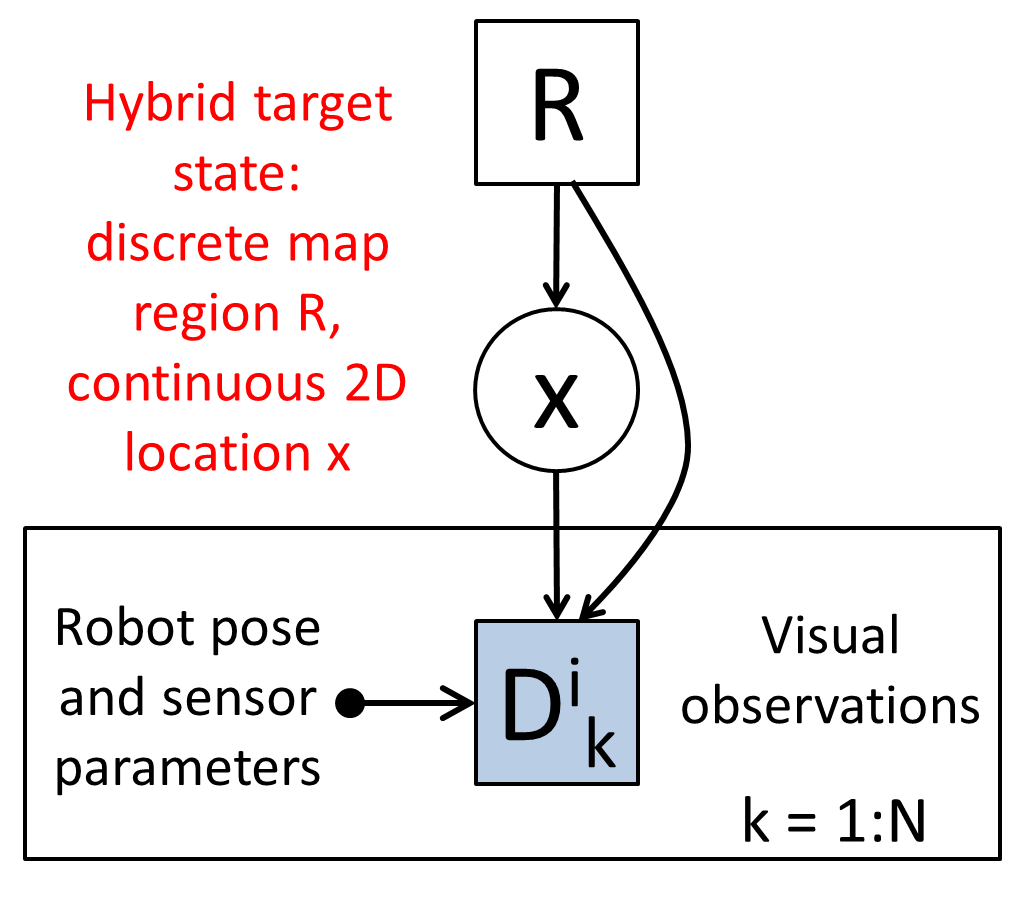} \\
\scriptsize (a) & \scriptsize (b) 
\end{tabular}
\caption{(a) Physical target search setup, showing $N_R=6$ discrete search regions over the 2D search space, (b) hybrid BN model used by each agent.
}
\label{fig:hybridBN}
\vspace{-0.5cm}
\end{figure}

%% file: searchResults.tex
\begin{figure*}[t]
\centering
\begin{tabular}{@{}c@{}c@{}c@{}}
\hspace*{-0.25 in}
\includegraphics[width=7.0cm]{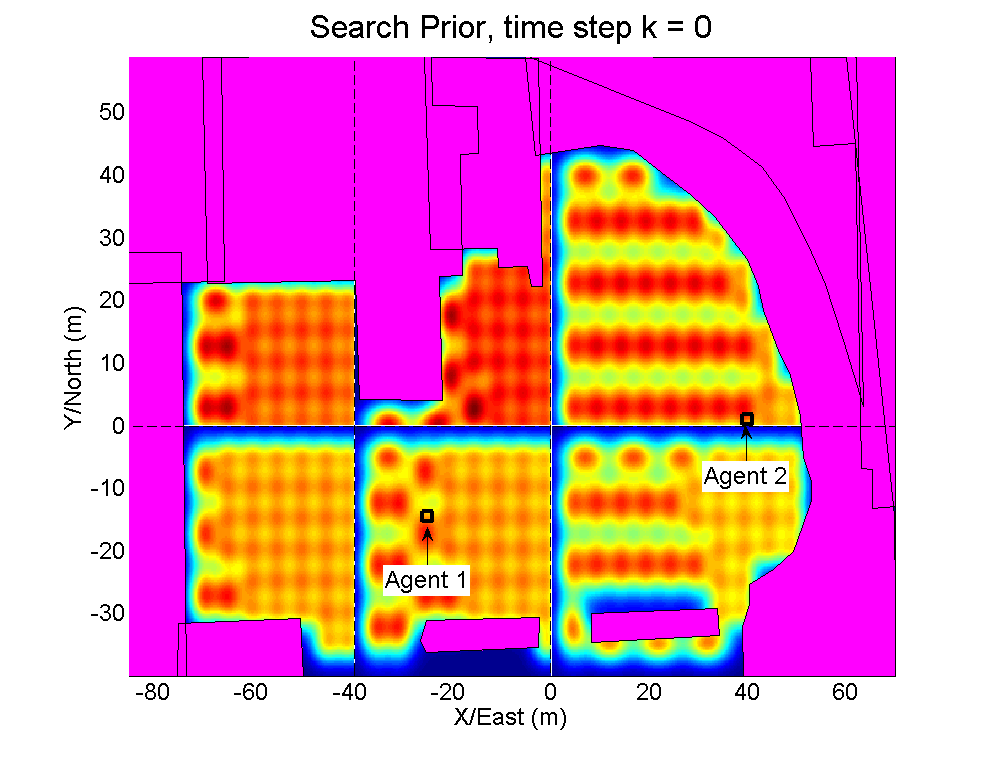} &
\hspace*{-0.35 in}
\includegraphics[width=7.0cm]{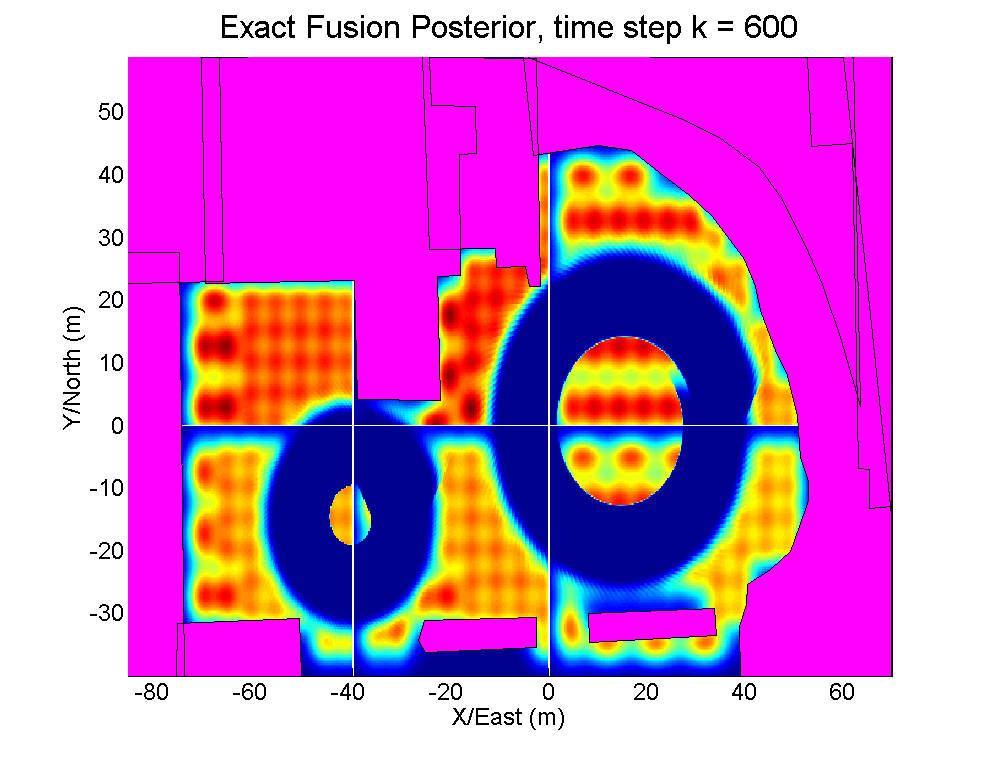} &
\hspace*{-0.35 in}
\includegraphics[width=7.0cm]{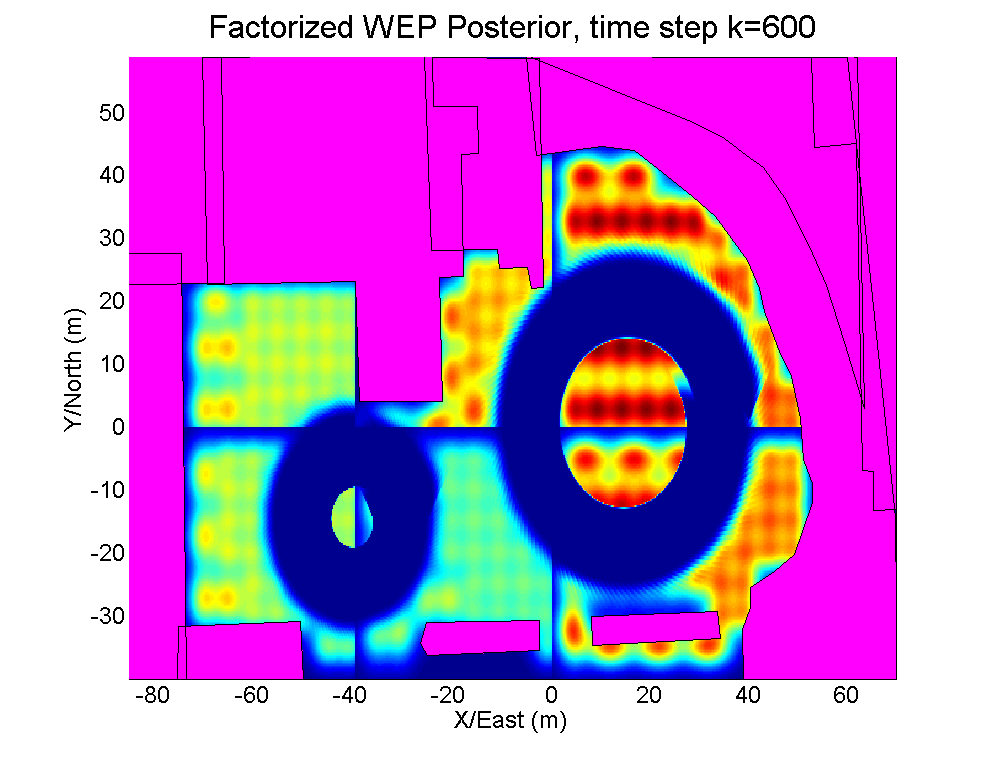} \\
\hspace*{-0.25 in}
\scriptsize (a) & \scriptsize (b) & \scriptsize (c) 
\end{tabular}
\caption{(a) Target search map prior distribution and initial locations of robot searchers, (b) exact DDF results after 600 time steps, (c) factorized WEP DDF results after 600 time steps, showing slight disagreement between $p_f(R)$ and $p_{f,\smC{WEP}}(R)$. Dark red/dark blue indicates high/low probability mass; magenta polygons show obstacle/boundary regions of search space.}
\label{fig:searchResults}
\vspace{-0.5cm}
\end{figure*}

%% file: omegaRsens.tex
\begin{figure}[tb]
	\centering
	\includegraphics[width=8.75cm]{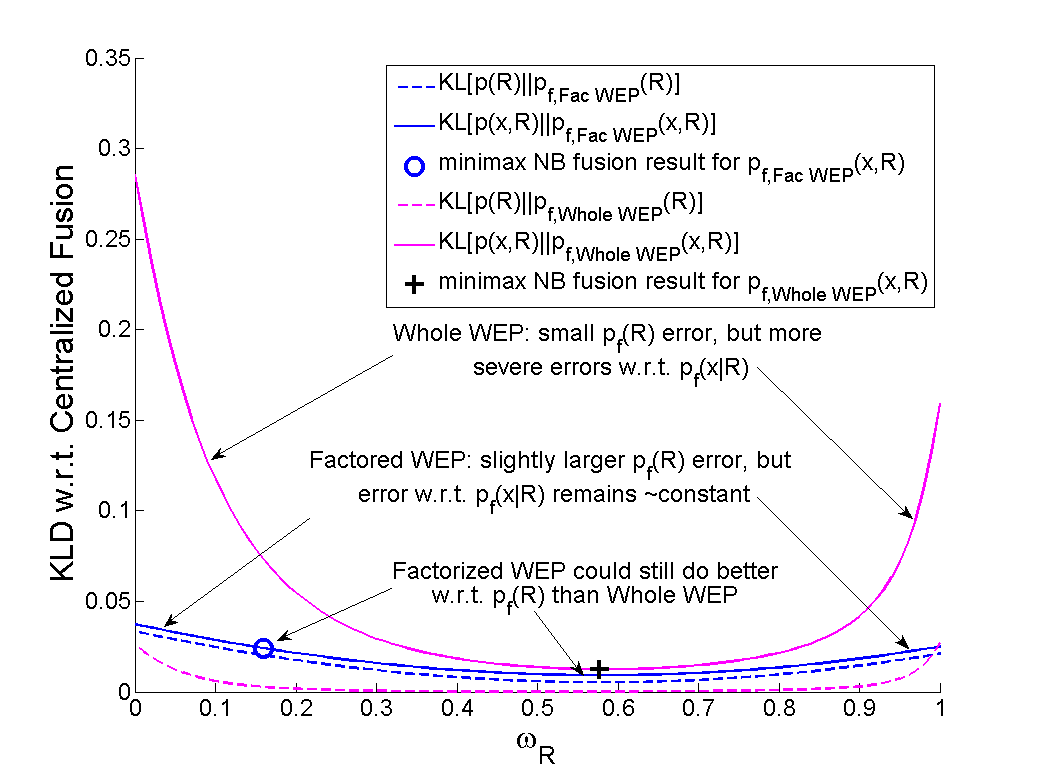}
	\caption{KLD losses for factorized and whole joint WEP DDF vs. $\omega_R$.}
	\vspace{-0.25 in}
	\label{fig:omegaRsens}
\end{figure}

%% file: mixtureddf.tex
This section describes our newly proposed mixture fusion algorithm, which overcomes the major limitations of other mixture fusion methods and produces accurate GM approximations to $\pexactddf$ and $\pwepddf$ (or any conditional factors thereof) for general GM DDF scenarios. Our technique is highly parallelizable and provides a unified approach to high fidelity recursive fusion of complex pdfs for both exact and WEP DDF. 

If we replace $p_i(x_k)$ and $p_j(x_k)$ with GM pdfs in either \pareqref{exactDef} or \pareqref{wepcommoninfo} and let $u(x_k)$ be the corresponding (estimated) non-Gaussian common information pdf (i.e. $p_c(x_k)$ or $\hat{p}_c(x_k)$), 
then this gives for exact DDF (and likewise for WEP DDF)
\begin{align}
\pexactddf 
\propto \frac{ \bigparenth{ \GaussianMix{i}{q}{M}{w}{\mu}{\Sigma}{x_k}} \bigparenth{ \GaussianMix{j}{r}{M}{w}{\mu}{\Sigma}{x_k} } }{ u(x_k)}, \nonumber
\end{align}
which is equivalent to
\begin{align}
\pexactddf \propto \sum_{q=1}^{M^i}{ \sum_{r=1}^{M^j}{ w^i_q w^j_r \frac{ {\cal N}(x_k; \mu^i_q, \Sigma^i_q) {\cal N}(x_k; \mu^j_r, \Sigma^j_r) }{ u(x_k) } } }. \label{eq:doublesum}
\end{align}
Using the fact that the product of two Gaussian pdfs is another unnormalized Gaussian pdf, this can be further simplified to
\begin{align}
\pexactddf \propto \sum_{q=1}^{M^i}{ \sum_{r=1}^{M^j}{ w^{ij}_{qr} \frac{ \bar{z}^{ij}_{qr} {\cal N}(x_k; \mu^{ij}_{qr}, \Sigma^{ij}_{qr})}{ u(x_k)} } },
\label{eq:ddfGMsum}
\end{align}
where each numerator term results from component-wise `Naive Bayes' fusion of $p_i(x_k)$ and $p_j(x_k)$,
\begin{align}
\Sigma^{ij}_{qr} &= \inv{ \bigsquare{ \inv{\bigparenth{\Sigma^{i}_{q}}} + \inv{\bigparenth{ \Sigma^{j}_{r}}} } }, \\
\mu^{ij}_{qr} &= \Sigma^{ij}_{qr} \bigsquare{ \inv{\bigparenth{\Sigma^{i}_{q}}}\mu^{i}_{q} + \inv{\bigparenth{ \Sigma^{j}_{r}}}\mu^{j}_{r} },  \\
\tilde{w}^{ij}_{qr} &= w^i_q w^j_r \bar{z}^{ij}_{qr}, \\
\bar{z}^{ij}_{qr} &= {\cal N}(\mu^{i}_{q}; \mu^{j}_{r}, \bigparenth{\Sigma^{i}_{q} + \Sigma^{j}_{r} }).
\end{align}
Eq.\pareqref{ddfGMsum} is thus a mixture of non-Gaussian components formed by the ratio of a single (unnormalized) Gaussian pdf and non-Gaussian pdf $u(x_k)$. Although not a normalized closed-form pdf, each component of \pareqref{ddfGMsum} tends to concentrate most of its mass around $\mu^{ij}_{qr}$. In particular, as $x_k$ moves away from $\mu^{ij}_{qr}$, the covariance $\Sigma^{ij}_{qr}$ (which is `smaller' than either $\Sigma^i_q$ or $\Sigma^j_r$) forces each Gaussian numerator term to decay more rapidly than $\frac{1}{u(x_k)}$ grows. This insight suggests that a good GM approximation to either $\pexactddf$ or $\pwepddf$ can be found by approximating each pdf ratio term in \pareqref{ddfGMsum} with a moment-matched Gaussian pdf, which leads to the GM approximation
\begin{align}
\pexactddf &\approx \frac{1}{\eta} \sum_{q=1}^{M^i}{ \sum_{r=1}^{M^j}{ \tilde{w}^{*}_{qr} {\cal N}(x_k; \mu^{*}_{qr}, \Sigma^{*}_{qr})} },
\label{eq:exactddfgmapprox} \\
\mbox{where \ } \ \tilde{w}^{*}_{qr} &= w^i_q w^j_r \cdot \EV{ 1 }_{p_{qr}(x_k)}, \label{eq:wstar} \\
\mu^{*}_{qr} &= \EV{ x_k } _{p_{qr}(x_k)}, \label{eq:mustar} \\
\Sigma^{*}_{qr} &= \EV{ x_k x^T_k }_{p_{qr}(x_k)} - \mu^{*}_{qr} (\mu^{*}_{qr})^T, \label{eq:sigstar} \\
\eta &= \sum_{q=1}^{M^i}{ \sum_{r=1}^{M^j}{ \tilde{w}^{*}_{qr} }}, \ \  
p_{qr}(x_k) \propto \frac{ \bar{z}^{ij}_{qr} {\cal N}(x_k; \mu^{ij}_{qr}, \Sigma^{ij}_{qr})}{u(x_k)}. \nonumber
\end{align}
While the required moments cannot be found analytically, they can be quickly estimated via Monte Carlo importance sampling (IS) \cite{Robert04Book}, which exploits the identity
\begin{align}
\EV{f(x_k)}_{p_{qr}(x_k)} &= \EV{ \frac{p_{qr}(x_k)}{h_{qr}(x_k)} f(x_k) }_{h_{qr}(x_k)} \nonumber \\
 &= \EV{ \theta(x_k) f(x_k) }_{h_{qr}(x_k)}, \nonumber
\end{align} 
where $f(x_k)$ is a given moment function and $h_{qr}(x_k)$ is a proposal pdf for each mixand $p_{qr}(x_k)$ that is easy to sample from, has a shape `close' to $p_{qr}(x_k)$, and has support on $x_k$ such that $p_{qr}(x_k)>0 \Rightarrow h_{qr}(x_k)>0$ (both $p_{qr}(x_k)$ and $h_{qr}(x_k)$ need only be known up to normalizing constants). Given a set of $N_s$ samples $\set{x^s_k}_{s=1}^{N_s} \sim h_{qr}(x_k)$, we obtain the sampling estimate 
\begin{align}
\EV{f(x_k)}_{p_{qr}(x_k)} &\approx \sum_{s=1}^{N_s}{ \theta(x^s_k) f(x^s_k) }, 
\ \ \theta(x^s_k) &\propto \frac{p_{qr}(x_k)}{h_{qr}(x_k)}. \nonumber 
\end{align}
Note that the moment calculations \pareqref{wstar}-\pareqref{sigstar} for each $qr$ term in \pareqref{exactddfgmapprox} can be easily parallelized.

There are many possible ways to select $h_{qr}(x_k)$ for each $p_{qr}(x_k)$; a particularly convenient (though not necessarily optimal) choice is $h_{qr}(x_k) = {\cal N}(x_k; \mu^{ij}_{qr}, \Sigma^{\smC{samp}}_{qr})$ for some suitable $\Sigma^{\smC{samp}}_{qr}$. This works well in practice as long as $h_{qr}(x_k)$ adequately covers the major support regions of $p_{qr}(x_k)$, i.e. if $(\Sigma^{\smC{samp}}_{qr} - \Sigma^{*}_{qr})$ is positive semi-definite and $p_{qr}(x_k)$ does not have too many widely separated modes. We have found that one effective strategy for low dimensional applications (i.e. $\leq$5 states) is to select
\begin{align}
\Sigma^{\smC{samp}}_{qr} = \arg \max(|\Sigma^{i}_q|, |\Sigma^{j}_r|, |\Sigma^{\smC{def}}|), \nonumber
\end{align}
where $\Sigma^{\smC{def}} = \alpha \cdot \eye$ and tuning parameter $\alpha$ represents a conservative upper bound on the expected variance for any posterior mixand in any dimension. This approach may not work well in large state spaces ($\geq$5 states) or for multimodal $p_{qr}(x_k)$. In future work, more sophisticated adaptive IS techniques \cite{Robert04Book} will be investigated for robust selection of $h_{qr}(x_k)$. 
\subsection{Illustrative 2D Example}
Figure \ref{fig:simple2Ddemo} shows a simple 2D example of WEP DDF for two GMs $p_i(x_k)$ and $p_j(x_k)$ using various mixture fusion approximations. Fig. \ref{fig:simple2Ddemo} (c) shows a high fidelity grid-based approximation to $\pwepddf$, which is very closely matched by the GM produced by our mixture fusion technique in Fig. \ref{fig:simple2Ddemo} (d) (the KL divergence between both pdfs is 0.0034 nats). Fig. \ref{fig:simple2Ddemo} (e) shows the fusion result obtained by the particle condensation method of \cite{Ahmed-RSS-12}, which relies on the EM algorithm to learn a GM approximation of $\pwepddf$. Although this captures the general shape of the true pdf, the parameter estimates are extremely sensitive to initial guesses and tend to get trapped at poor local solutions, which leads to greater information loss (KLD of 0.1035 nats). Fig. \ref{fig:simple2Ddemo} (f) shows that the fusion GM produced by first order covariance intersection (FOCI) \cite{Julier06} loses even more information (KLD of 0.6972 nats), due to its overly conservative nature. 

Note that the results in Fig. \ref{fig:simple2Ddemo} (d) and (e) are subject to variance from Monte Carlo IS; although not shown here, this variance is substantially lower for the proposed mixture fusion method due to the fact that IS is applied to each non-Gaussian mixand of $\pwepddf$ individually, rather than to $\pwepddf$ as a whole. Furthermore, the proposed method does not require the solution to a nonlinear optimization problem as in the particle condensation method. Like the FOCI approximation, the proposed technique also automatically leads to a larger but still finite number of mixands $M^f = M^iM^j$ in the GM approximation. Post-hoc GM compression methods can be used to control $M^f$ in real applications while minimizing information loss, although these typically have $O((M^f)^2)$ or $O((M^f)^3)$ memory and time costs. In future work, we will investigate ways to control $M^f$ `on the fly', e.g. by culling/merging $qr$ terms with very small weights in \pareqref{exactddfgmapprox} before/during IS moment-matching calculations.
%
\input{combinedSimple2Dexample}
%

%% file: combinedSimple2Dexample.tex
\begin{figure}[t]
\centering
\begin{tabular}{@{}c@{}c@{}}
\hspace*{-0.25 in}
\includegraphics[width=4.95cm]{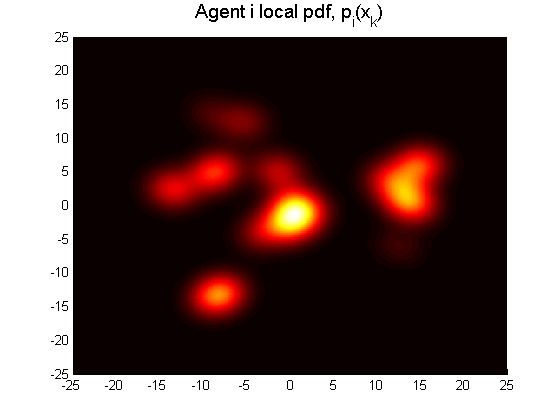} &
\hspace*{-0.15 in}
\includegraphics[width=4.95cm]{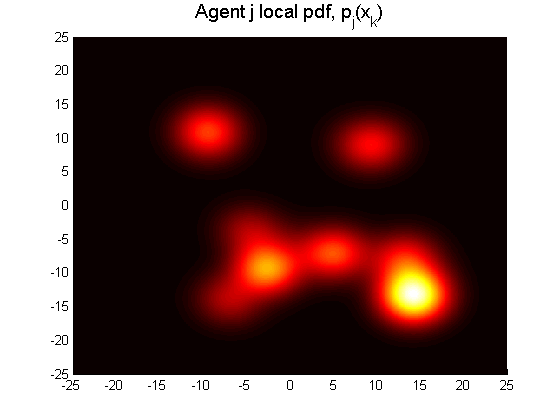} \\
\hspace*{-0.25 in}
\scriptsize (a) & \scriptsize (b) \\
\hspace*{-0.25 in}
\includegraphics[width=4.95cm]{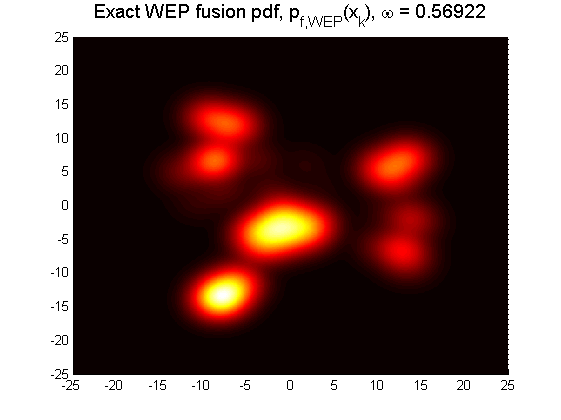} &
\hspace*{-0.15 in}
\includegraphics[width=4.95cm]{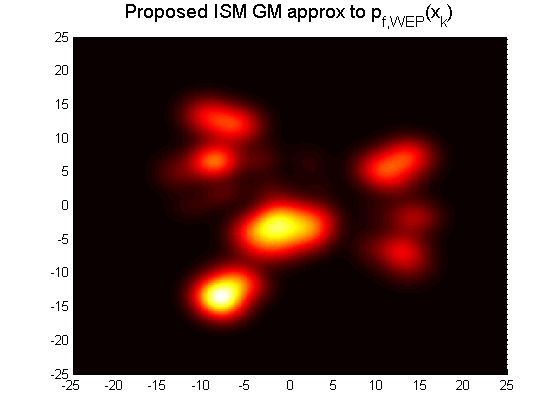} \\
\hspace*{-0.25 in}
\scriptsize (c) &  \scriptsize (d) \\
\hspace*{-0.25 in}
\includegraphics[width=4.95cm]{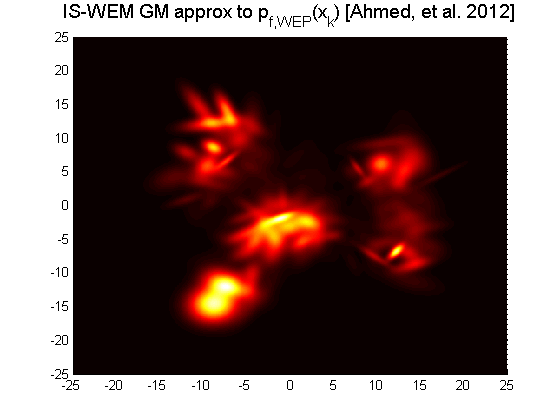} &
\hspace*{-0.15 in}
\includegraphics[width=4.95cm]{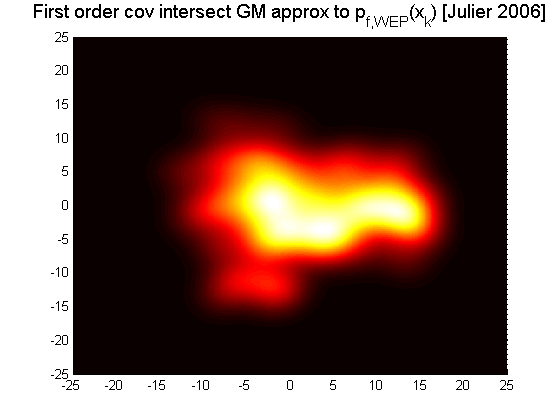} \\
\hspace*{-0.25 in}
\scriptsize (e) &  \scriptsize (f)
\end{tabular}
\caption{(a)-(b) GMs $p_i(x)$ and $p_j(x)$ with $M_i=M_j=14$; (c) ground truth grid-based $\pwepddf$ ($\omega = 0.56922$); (d) proposed GM approximation; (e) weighted EM GM approximation; (f) FOCI GM approximation.
}
\label{fig:simple2Ddemo}
\vspace{-0.5cm}
\end{figure}